\documentclass[10pt,twocolumn,letterpaper]{article}

\usepackage[pagenumbers]{cvpr} 

\usepackage[dvipsnames]{xcolor}

\def \customparskip {.3em}
\renewcommand{\paragraph}[1]{\vspace{\customparskip}\noindent\textbf{#1}}

\definecolor{turquoise}{cmyk}{0.65,0,0.1,0.3}
\definecolor{purple}{rgb}{0.65,0,0.65}
\definecolor{dark_green}{rgb}{0, 0.5, 0}
\definecolor{orange}{rgb}{0.8, 0.6, 0.2}
\definecolor{red}{rgb}{0.8, 0.2, 0.2}
\definecolor{darkred}{rgb}{0.6, 0.1, 0.05}
\definecolor{blueish}{rgb}{0.0, 0.3, .6}
\definecolor{light_gray}{rgb}{0.7, 0.7, .7}
\definecolor{pink}{rgb}{0.9, 0, 0.6}
\definecolor{greyblue}{rgb}{0.25, 0.25, 1}
\definecolor{teal}{rgb}{0.0, 0.4, 0.4}

\newcommand{\hparam}[1]{\lambda_{\text{#1}}}

\newcommand{\image}{{\mathbf{x}}}
\newcommand{\latent}{{\mathbf{z}}}

\newcommand{\numstep}{{T}}
\newcommand{\encoder}{{\mathcal{E}}}
\newcommand{\decoder}{{\mathcal{D}}}
\newcommand{\denoiser}[1]{{\epsilon}_\text{#1}}
\newcommand{\viewembedder}{{\mathcal{V}}}

\newcommand{\view}{{\mathbf{v}}}
\newcommand{\sample}{{\Phi}}
\newcommand{\viewidx}{{f}}
\newcommand{\numframe}{{F}}

\renewcommand{\paragraph}[1]{\smallskip\noindent\textbf{#1}}

\definecolor{cvprblue}{rgb}{0.21,0.49,0.74}
\usepackage[pagebackref,breaklinks,colorlinks,citecolor=cvprblue]
{hyperref}
\usepackage{gensymb}
\def\paperID{2582} 
\def\confName{CVPR}
\def\confYear{2024}

\title{ViVid-1-to-3: Novel \underline{Vi}ew Synthesis with  \underline{Vid}eo Diffusion Models}

\author{
  Jeong-gi Kwak\textsuperscript{1,2}\hspace{0.5mm}\thanks{Equal contribution}\hspace{1.5mm}\thanks{Work done while visiting University of British Columbia} \hspace{3mm}
  Erqun Dong\textsuperscript{1,3,4}\hspace{0.5mm}\footnotemark[1]\hspace{1.5mm}\footnotemark[2] \hspace{3mm}
  Yuhe Jin\textsuperscript{1} \hspace{3mm}
  Hanseok Ko\textsuperscript{2} \hspace{3mm} 
  \\
  Shweta Mahajan\textsuperscript{1, 5} \hspace{6mm}
  Kwang Moo Yi\textsuperscript{1, 4}\hspace{0.5mm}\footnotemark[2] 
  \\ 
  \textsuperscript{1} University of British Columbia \hspace{3mm}
  \textsuperscript{2} Korea University \hspace{3mm}
  \textsuperscript{3} McGill University
  \\
  \textsuperscript{4} Haiper Ltd. \hspace{6mm}
  \textsuperscript{5} Vector Institute for AI
  \\{\small {\url{https://jgkwak95.github.io/ViVid-1-to-3/}}}
}

\makeatletter
\apptocmd\@maketitle{{\teaserfigure{}\par}}{}{}
\makeatother

\def\myshift#1{\raisebox{0.5ex}}
\newcommand{\teasernobox}{
\begin{subfigure}[b]{\linewidth}
    \centering
 	\includegraphics[width=\linewidth]{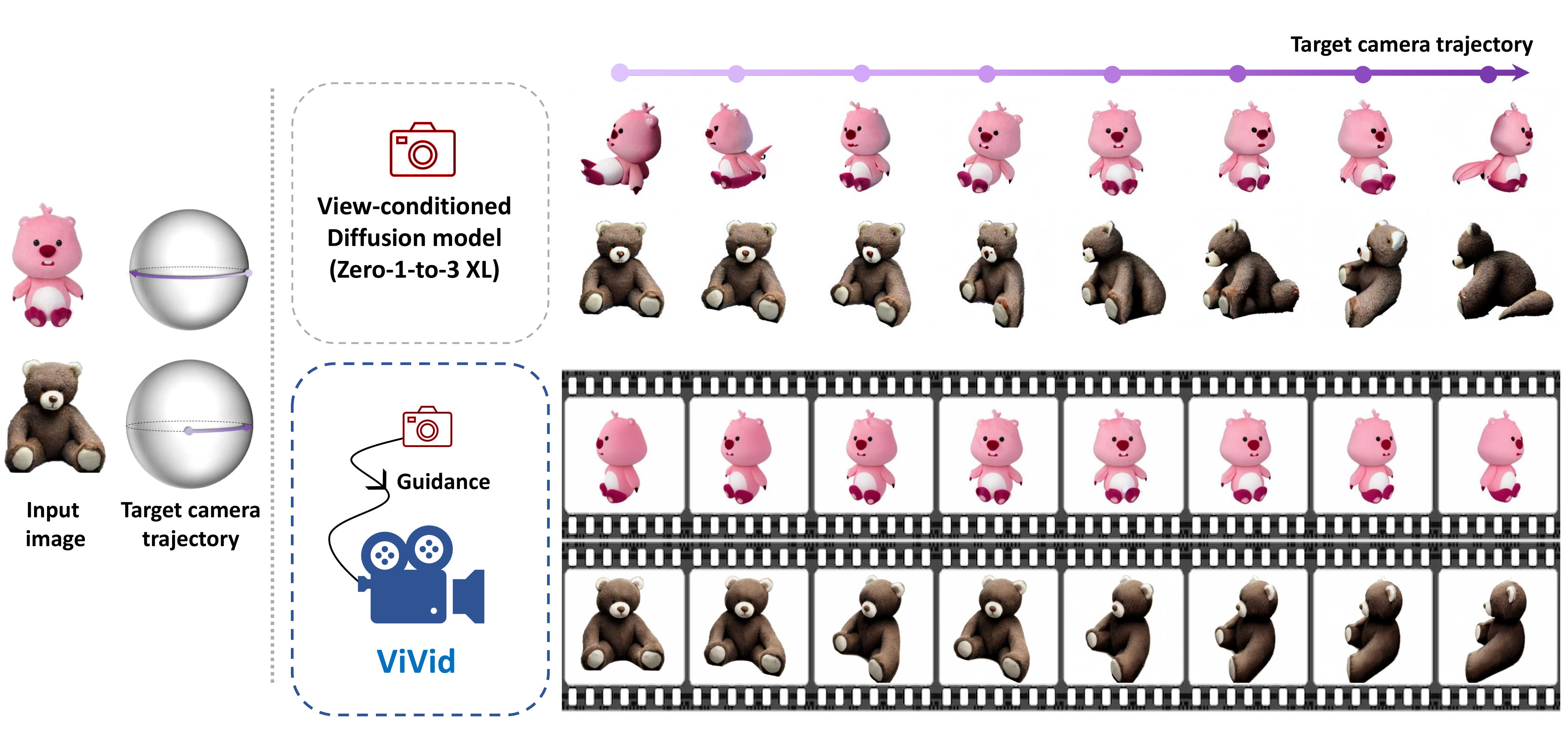}
\end{subfigure}
}

\newcommand{\teaserfigure}{
\vspace{-5mm}
\captionsetup{type=figure}

\teasernobox

\vspace{-2mm}
\setcounter{figure}{0} 
\captionsetup{type=figure}
\captionof{figure}{
{\bf Teaser} --
we present a strikingly simple training-free method to make already available novel-view synthesis diffusion models more consistent both in terms of the desired viewing angle and the content---combining it with video diffusion.
As shown in the example, the results of our method are more consistent with the input images and correspond more to the target views.
}
\vspace{4mm}
\label{fig:teaser}
}
\vspace{-1em}

\begin{document}
\maketitle\begin{abstract}
Generating novel views of an object from a single image is a challenging task. It requires an understanding of the underlying 3D structure of the object from an image and rendering high-quality, spatially consistent new views. While recent methods for view synthesis based on diffusion have shown great progress, achieving consistency among various view estimates and at the same time abiding by the desired camera pose remains a critical problem yet to be solved. In this work, we demonstrate a strikingly simple method, where we utilize a pre-trained video diffusion model to solve this problem. Our key idea is that synthesizing a novel view could be reformulated as synthesizing a video of a camera going around the object of interest---a scanning video---which then allows us to leverage the powerful priors that a video diffusion model would have learned. Thus, to perform novel-view synthesis, we create a smooth camera trajectory to the target view that we wish to render, and denoise using both a view-conditioned diffusion model and a video diffusion model. By doing so, we obtain a highly consistent novel view synthesis, outperforming the state of the art. 

\end{abstract}

\section{Introduction}
\label{sec:intro}

Novel view synthesis from a single image is an interesting problem in Computer Vision as it requires an understanding of the 3D characteristics of an object or a scene, simply by looking at a 2D image.
Recent methods have utilized Neural Radiance Fields (NeRF)~\cite{YuYTK21,jain2021putting}, and more recently diffusion models~\cite{watson2022novel,abs-2304-02602}, including those that expand upon Stable Diffusion~\cite{rombach2022high} such as Zero-1-to-3~\cite{liu2023zero1to3}.
While recent methods show high-quality rendering from a given view,
when examined more carefully, these models lack consistency~\cite{liu2023zero1to3}, and their poses are also not accurate.
This is because diffusion models are mainly geared towards generating images that look good. 
Moreover, diffusion models learn the 3D consistency implicitly, not explicitly.
This limitation of content and view consistency is not resolved even when models are trained of large datasets~\cite{deitke2023objaverse,deitke2023objaversexl}; see~\cref{fig:teaser}.

\begin{figure}
    \centering
    \includegraphics[width=\linewidth]{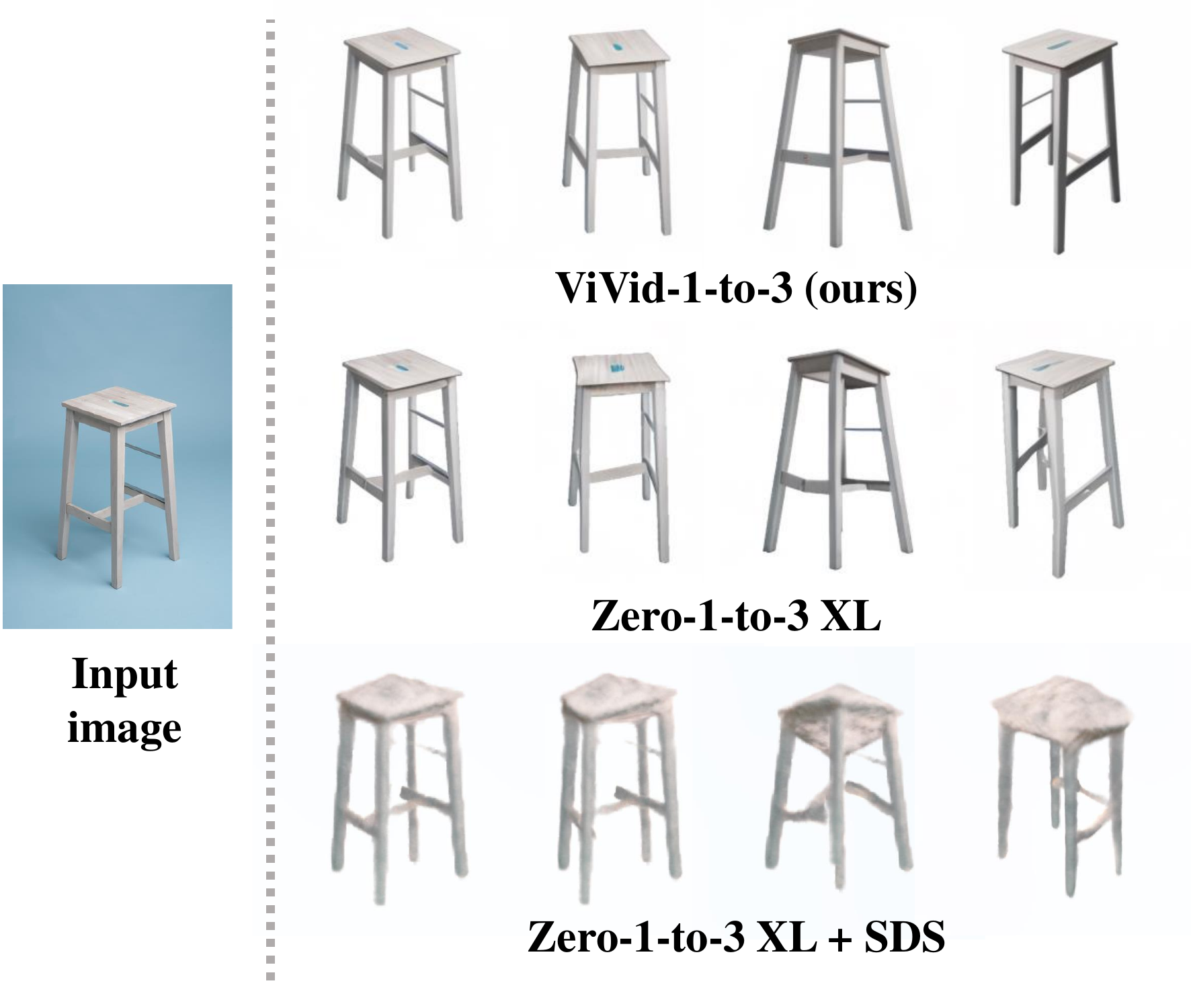}
    \caption{
    {\bf Example rendering inconsistencies -- }
    we show example novel views generated for a chair image as shown on the left.
    Our method provides improved consistency with the input image, compared to both Zero-1-to-3 XL~\cite{liu2023zero1to3,deitke2023objaversexl}, a pure 2D novel-view diffusion model, and even when it is combined with the Score Distillation Sampling (SDS)~\cite{poole2022dreamfusion} for improved 3D consistency.
    Note that the 3D distilled version exhibits blurriness due to the pose errors that Zero-1-to-3 XL makes.
     }
    \label{fig:2d_vs_3d}
\end{figure}

Very recently, researchers have thus focused their efforts on making diffusion model-based novel view synthesis more `consistent'.
These include methods that train or finetune 2D diffusion models to make them more consistent~\cite{liu2023syncdreamer,weng2023consistent123,yang2023consistnet,shi2023mvdream,shi2023zero123plus,deitke2023objaversexl} or embed 3D constraints in the form of 3D representations~\cite{MildenhallSTBRN20,wang2021neus,kerbl20233d} through score distillation techniques~\cite{poole2022dreamfusion,WangDLYS23,wang2023prolificdreamer,abs-2306-17843,tang2023dreamgaussian}.
While they provide improved results, there are still several limitations.   
2D view-conditioned models often require retraining~\cite{liu2023syncdreamer,weng2023consistent123,shi2023zero123plus} which is costly, and as shown in \cref{fig:2d_vs_3d} (middle), even with this additional training they still have considerable room for improvement.
In the case of methods that bring in explicit 3D representations such as NeRF~\cite{MildenhallSTBRN20} or Gaussian Splatting~\cite{kerbl20233d} on top~\cite{liu2023one,abs-2306-17843,tang2023dreamgaussian}, suffer from poor visual quality often due to the blurs that the pose inconsistencies of 2D models bring; see~\cref{fig:2d_vs_3d} (bottom). 

In this work, we show that a strikingly simple solution that \emph{does not require any new training or fine-tuning} exists for this problem---leveraging pre-trained video diffusion models.
Our key idea is that, given their recent improvement in quality, video diffusion models~\cite{zeroscope,guo2023animatediff} can be used as strong priors for any task that can be represented as a video---including the current task of novel view synthesis.
For example, consider generating a video of someone scanning an object. 
If we were to generate such a  video, where the first frame of the video corresponds to the object that we wish to generate a novel view of, and the last frame is from the target novel view, generating this video is effectively equivalent to the single image novel view synthesis task, except now with redundant intermediate frames. 
Importantly, we already have pre-trained public models for both generating these individual frames---view-conditioned diffusion models~\cite{liu2023zero1to3}---and the video as a whole---video diffusion models~\cite{zeroscope,guo2023animatediff}.


To implement our method we use Zero-1-to-3 XL~\cite{liu2023zero1to3,deitke2023objaversexl} and ZeroScope~\cite{zeroscope}, both of which are based on Stable Diffusion~\cite{rombach2022high}.
We generate a smooth camera trajectory from the frontal view to the desired view and provide Zero-1-to-3 XL with the individual camera locations for each frame, along with the input image.
We then denoise a video, where the noise estimates for each frame are a combination of noise estimates from both diffusion models.
We investigate multiple different strategies for combining the noise estimates, and find that starting to denoise with equal emphasis on both diffusion models and then reducing the weight on the video diffusion model to half throughout the denoising process is the best.

To evaluate our method we rely on 100 shapes from the Google Scanned Object (GSO) dataset~\cite{downs2022google}.
As we are able to render these shapes from any view, we compare the novel-view synthesis results with the ground-truth rendering.
We find that, due to the difficulty of the novel-view synthesis tasks, it is very easy to have \emph{minor misalignments}, and the typical image quality metrics do not provide a complete view.
We thus propose a novel metric based on optical flow, and measure the \emph{spatial deviations} of the novel-view renderings.
Our metric, together with the standard ones provides a holistic view of the performances of each method.





\section{Related work}
\label{sec:related}


We first review novel-view synthesis methods based on explicit geometric constraints, then discuss the more recent trend of using diffusion models.
As we rely on video diffusion, we also briefly discuss noteworthy works.

\paragraph{Novel view synthesis with geometric constraints.}
Early work on novel view synthesis recovers the 3D structure of a scene by incorporating geometric prior such as camera parameters \cite{SnavelySS06, SchonbergerF16}. 
Since then, as in many other areas in computer vision, deep learning-based methods have been proposed~\cite{RieglerK20,RieglerK21,freer2022novel}, which often combine traditional 3D geometry aware multi-view synthesis~\cite{AgarwalSSSS09,GoeseleSCHS07} with modern deep learning.
Other works leverage voxels~\cite{SitzmannTHNWZ19,LombardiSSSLS19}, depth maps~\cite{FlynnNPS16,TuckerS20} or epipolar constraints~\cite{LandreauT22} in their model.

More recently, since the introduction of Neural Radiance Fields (NeRF)~\cite{MildenhallSTBRN20}, 3D constraints via volume rendering have become popular.
While training a NeRF requires multiple views, PixelNeRF~\cite{YuYTK21} utilizes convolutional features of pre-trained deep networks~\cite{he2016deep} to reduce the number of required views, as little as a single view.
GRF~\cite{Trevithick021} extends this idea to the concept of canonical space.
GenNVS~\cite{abs-2304-02602} further appends diffusion models into the pipeline for improved rendering.
While they have shown impressive progress, these methods require per-dataset training, which limits their applicability.

\paragraph{Novel-view synthesis with diffusion models.}
Recently, like many other applications that involve image generation~\cite{rombach2022high,meng2021sdedit,hedlin2023unsupervised,khani2023slime}, there has been a flurry of research for novel view synthesis based on diffusion models.
These include those that directly aim to generate 2D novel views conditioned on the input image and the camera pose~\cite{watson2022novel,liu2023zero1to3}.
Among them, Zero-1-to-3~\cite{liu2023zero1to3}, being based on Stable Diffusion~\cite{rombach2022high}, has demonstrated impressive results, benefiting from the original weights of Stable Diffusion that have been trained with a very large dataset~\cite{schuhmann2022laion}.
This method has been further trained on a large-scale 3D dataset~\cite{deitke2023objaverse,deitke2023objaversexl}, further improving its performance---we use this model.

While these direct 2D methods have shown impressive rendering quality, as shown in \cref{fig:2d_vs_3d}, they often fall short when it comes to the consistency of what they render, and also from where they are viewed.
Various approaches, concurrently to our work~\cite{liu2023syncdreamer,shi2023mvdream,weng2023consistent123,yang2023consistnet,shi2023zero123plus} have thus been presented in an attempt to make them more `consistent'.
They, however, require re-training or fine-tuning, and sometimes only provide restricted views~\cite{liu2023syncdreamer}.
Our method suffers from neither of these shortcomings.

Alternatively, to enforce consistency, methods that aim to \emph{distill} what diffusion models have learned to 3D representations have also been suggested.
DreamFusion~\cite{poole2022dreamfusion} and Score Jacobian Chaining (SJC)~\cite{WangDLYS23} leverage a pre-trained diffusion model~\cite{SahariaCSLWDGLA22} to distill it into a NeRF representation~\cite{MildenhallSTBRN20}, allowing text-to-3D, which can also be utilized for novel-view synthesis.
Various followups~\cite{Lin0TTZHKF0L23,Melas-KyriaziL023,Xu0WFWW23,abs-2303-07937,JainMBAP22,Chen_2023_ICCV,wang2023prolificdreamer,armandpour2023re,XuW0CSQG23} have since then been suggested to improve the rendering quality via additional constraints such as subject-driven diffusion guidance~\cite{abs-2303-13508} and frontal camera position~\cite{Lin0TTZHKF0L23,abs-2303-07937}.
%

More recent, concurrent works~\cite{liu2023one,abs-2306-17843,long2023wonder3d} have further integrated this idea with view-conditioned diffusion methods~\cite{liu2023zero1to3}, but due to misalignments between desired camera pose and the outcome of 2D part of their pipelines, their results can become blurry (\cref{fig:2d_vs_3d}).
Our method is complementary to these efforts in that we still rely on a pure 2D pipeline, yet allow more consistent renderings.

\paragraph{Video diffusion models.} 
We also briefly review video diffusion models since we utilize them.
A recent trend in video diffusion models is to take advantage of the advancements in image diffusion models by factorizing space and time~\cite{HoSGC0F22,abs-2210-02303,SingerPH00ZHYAG23,abs-2211-11018,BlattmannRLD0FK23,voleti2022mcvd}.
They include methods that use cascaded diffusion models~\cite{abs-2210-02303}, and spatio-temporal interpolation~\cite{SingerPH00ZHYAG23}.
Among them, methods that extend already trained test-to-image latent diffusion models such as Stable Diffusion~\cite{rombach2022high} have become popular.
For example, \citet{BlattmannRLD0FK23} choose to train separate temporal layers, \citet{guo2023animatediff} injects motion modules, ZeroScope~\cite{zeroscope} finetunes VideoFusion~\cite{VideoFusion}, which decomposes the diffusion process as a sum of a base noise and a residual noise.
Among them, we use ZeroScope, as their models are publicly available, and they also build on top of Stable Diffusion, as is the case of view-synthesis diffusion model Zero-1-to-3 XL~\cite{liu2023zero1to3,deitke2023objaversexl}.

\begin{figure*}
    \centering
    \includegraphics[width=\linewidth]{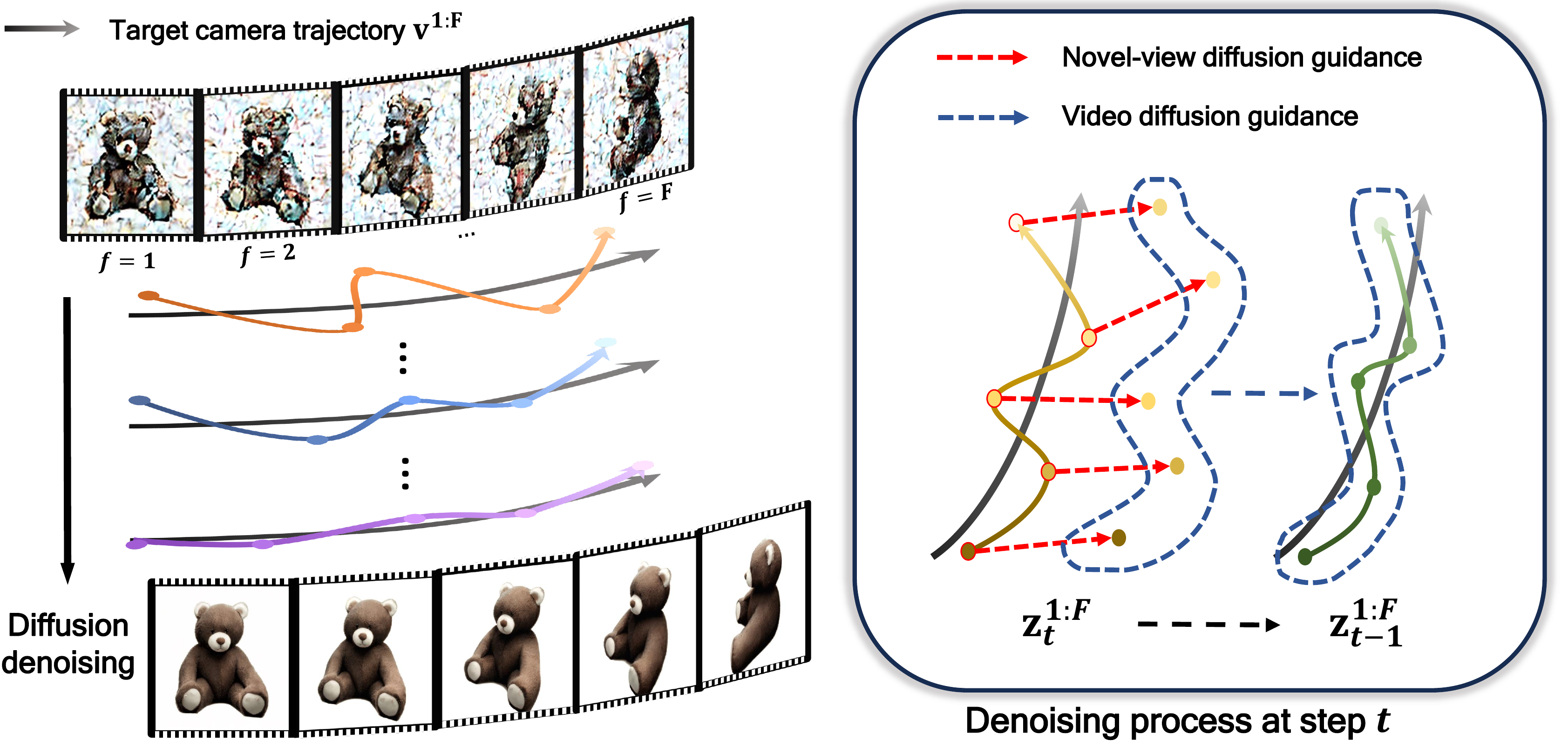}
    \caption{
        {\bf Overview -- }
        to synthesize a target novel view of a given object, we synthesize along a smooth camera trajectory, starting from an initial view and ending at the target.
        We then denoise to generate images with guidance from two diffusion models: a novel-view synthesis diffusion model; and a video diffusion model.
        The video diffusion model helps create a smooth camera trajectory and preserve consistency, which the novel-view synthesis diffusion model can lack.
        The two together, as shown, provide high-quality novel-view synthesis, that is consistent in both the content and the desired camera views.
    }
    \label{fig:overview}
\end{figure*}

\section{Method}
\label{sec:method}

The key idea behind our method is strikingly simple---video diffusion models can be used as a strong prior in conjunction with novel-view diffusion models to improve consistency in synthesized views, \emph{without} any training or fine-tuning.
To do this, instead of directly generating a novel view, we propose to generate a camera trajectory towards the camera view that we wish to render.

As illustrated in \cref{fig:overview}, images generated by the view-conditioned diffusion model for generating novel views often do not faithfully follow the target camera path, and can also have (object) pose errors.
As such renderings, when viewed as a video, would not look like a natural video due to abrupt motions, using a video diffusion model helps to smooth out this error.
This results in renderings that abide by the requested novel pose, as well as the contents being consistent with the input view.

To explain our method, we first provide a brief overview of diffusion models to introduce the notations that we will use (\cref{sec:prelim}) then detail our method (\cref{sec:main_method}).

\subsection{Preliminary: diffusion models}
\label{sec:prelim}

Specifically among the diffusion models, we rely on Latent Diffusion Models (LDM)~\cite{rombach2022high,van2017neural,esser2021taming}.
Latent diffusion models utilize an encoder-decoder, $\encoder$-$\decoder$, pair---often pre-trained, \eg, with Vector-Quantized Generative Adversarial Nets (VQ-GAN)~\cite{rombach2022high}---to convert an input image $\image$ into a latent code $\latent=\encoder(\image)$ and transform it into noise by iteratively adding Gaussian noise for $\numstep$ steps~\cite{rombach2022high}.
The diffusion model then learns to estimate the amount of noise at a given time step $t$, which is then used to reverse the diffusion process and denoise to the corresponding signal $\latent$.
Specifically, denoting the denoising model as $\denoiser{}(\cdot)$, to generate an image we write
\begin{equation}
    \latent_{t-1} = \sample\left(
        \latent_t, \denoiser{}\left(\latent_t, t, y\right)
    \right)
    \label{eq:diffusion}
    ,
\end{equation}
where $\sample(\cdot)$ is an update (sampling) rule for denoising such as Denoising Diffusion Probabilistic Models (DDPM)~\cite{ho2020denoising},
and $y$ is the conditioning text.
Note that for clarity in notation, we drop text conditioning from \cref{eq:diffusion}.

We now formalize the two types of denoising diffusion models that we use, a model for novel-view synthesis, and one for generating videos.

\paragraph{Diffusion models for novel view synthesis.}
For novel view synthesis with diffusion models, 
instead of the text conditioning via $y$, it is often replaced with an encoding of the image and the desired camera pose~\cite{liu2023zero1to3,shi2023mvdream}.
Thus, denoising with this model takes the form,
\begin{equation}
    \latent_{t-1} = \sample\left(
        \latent_t, \denoiser{view}\left(\latent_t, t, \viewembedder(\image^0,\view)\right)
    \right)
    ,
    \label{eq:view_diffusion}
\end{equation}
where $\image^0$ is the input image, $\view$ denotes the camera poses to synthesize, and $\viewembedder$ is a mapping that encodes the input image $\image^0$ and the desired view $\view$ to a conditioning signal.

\paragraph{Video diffusion models.} 
While video diffusion models vary in their architectural designs and their sophisticated training schemes~\cite{BlattmannRLD0FK23,guo2023animatediff,VideoFusion,zeroscope}, at a high level they diffuse \emph{multiple} images (frames) together so that they form a video.
At any timestep of the denoising process, the denoising model is applied across all the frames, given by,
\begin{equation} 
    \latent_{t-1}^{1:\numframe} = \sample\left(
        \latent_t^{1:\numframe}, \denoiser{video}\left(\latent_t^{1:\numframe}, t, y\right)
    \right)
    ,
    \label{eq:video_diffusion}
\end{equation}
where $\latent^{1:\numframe}{=}\{\latent^1_t,\dots,\latent^\numframe_t\}$ denotes the $F$ frames of a video at a denoising step $t$.

\subsection{Video diffusion for novel view synthesis}
\label{sec:main_method}

To combine the two diffusion models for novel view synthesis, we generate a trajectory of views $\view^{1:\numframe}=\{\view^1,\dots,\view^\numframe\}$ where $\view^\numframe$ is the desired novel view.
We generate these views by creating a smooth trajectory of views through Spherical Linear Interpolation (Slerp)~\cite{dam1998quaternions}, starting from the initial view $\view^1$ to the target view $\view^F$.
For convenience, we set the initial view to be the same as $\view^0$ corresponding to the input image $\image^0$ (\ie, $\view^1:=\view^0$), but it can be any arbitrary view.
We then initialize the latent for each view $\latent_T^{1:F}$ by drawing them from a Gaussian distribution, that is, $\latent_T^{1:F}\sim\mathcal{N}(\mathbf{0},\mathbf{1})$.
With these, we iteratively denoise them according to the typical diffusion process
with the two denoisers $\denoiser{view}$ and $\denoiser{video}$ together.

Specifically, simplifying the notation for the denoiser estimates in \cref{eq:view_diffusion,eq:video_diffusion} by dropping $t$,  $y$, and $\viewembedder$, as $\denoiser{view}(\latent_t,\image^0,\view)$ and $\denoiser{video}(\latent^{1:\numframe})$,
we write our denoising process as
\begin{equation} \label{eq:our_denoising}
    \latent_{t-1}^{1:\numframe}=\sample\left(
        \latent_t^{1:\numframe}, \denoiser{both}^{1:\numframe}
    \right)
    ,
\end{equation}
where, denoting the view index as superscript $\viewidx$,
\begin{equation}
        \denoiser{both}^\viewidx
        =
        \hparam{view}
        \denoiser{view}(\latent^\viewidx,\image^0,\view^\viewidx)
        +
        \hparam{video}
        \denoiser{video}\left(\latent^{1:\numframe}\right)^\viewidx
        ,
        \label{eq:both}
\end{equation}
where $\hparam{view}$ and $\hparam{video}$ denote the hyperparameters controlling the influence of each denoising model.

Note that combining two noise estimates together is similar to how conditional and unconditional models are used together for Classifier Free Guidance~\cite{ho2022classifier}.
Here, we are combining two models, each with different conditioning and architecture to perform the \emph{same task} of generating the \emph{video} frames of the camera going about a specific trajectory.
Ultimately, for the generated frames to be precise, they must satisfy both models, which is the core idea that provides improved synthesis results as shown in \cref{fig:overview}.



\paragraph{Null prompting.}
As shown in \cref{eq:both}, $\denoiser{view}(\cdot)$, the denoiser for the novel views is already conditioned on the input image $\image^0$.
Because of this, there is no need for the user to provide the text prompt $y$ for $\denoiser{video}(\cdot)$.
We thus set it to a null prompt, that is, $y=\O$.
Hence, the guidance of the content for the video diffusion model is purely reliant on $\denoiser{view}(\cdot)$, the novel-view synthesis diffusion model.



\paragraph{Scheduling influence of each model.}
We found that the way the two noise estimates are combined in \cref{eq:both} has an impact on the behaviour of our method.
Too strong emphasis---large $\hparam{video}$---on the $\denoiser{video}(\cdot)$ estimates in later stages of the denoising process causes the entire generation to be overly smooth, while having too small of an impact---small $\hparam{video}$---in early stages restrict the impact of the video diffusion model, 
as the early stages determine the global structures~\cite{hertz2022prompt,balaji2022ediffi} often related to the camera view.
We thus propose a linearly decaying setup, where $\hparam{view}{=}1$ is set to a constant, and $\hparam{video}$ decay linearly from 1.0 at the 0-th step to 0.5 at the 50-th step.
We ablate our choice in \cref{sec:ablation}.

\section{Results}
\label{sec:result}

\subsection{Datasets and experimental setup}

We will release the code and the experimental settings to make our results completely reproducible.

\begin{figure}
    \centering
    \includegraphics[width=\linewidth]{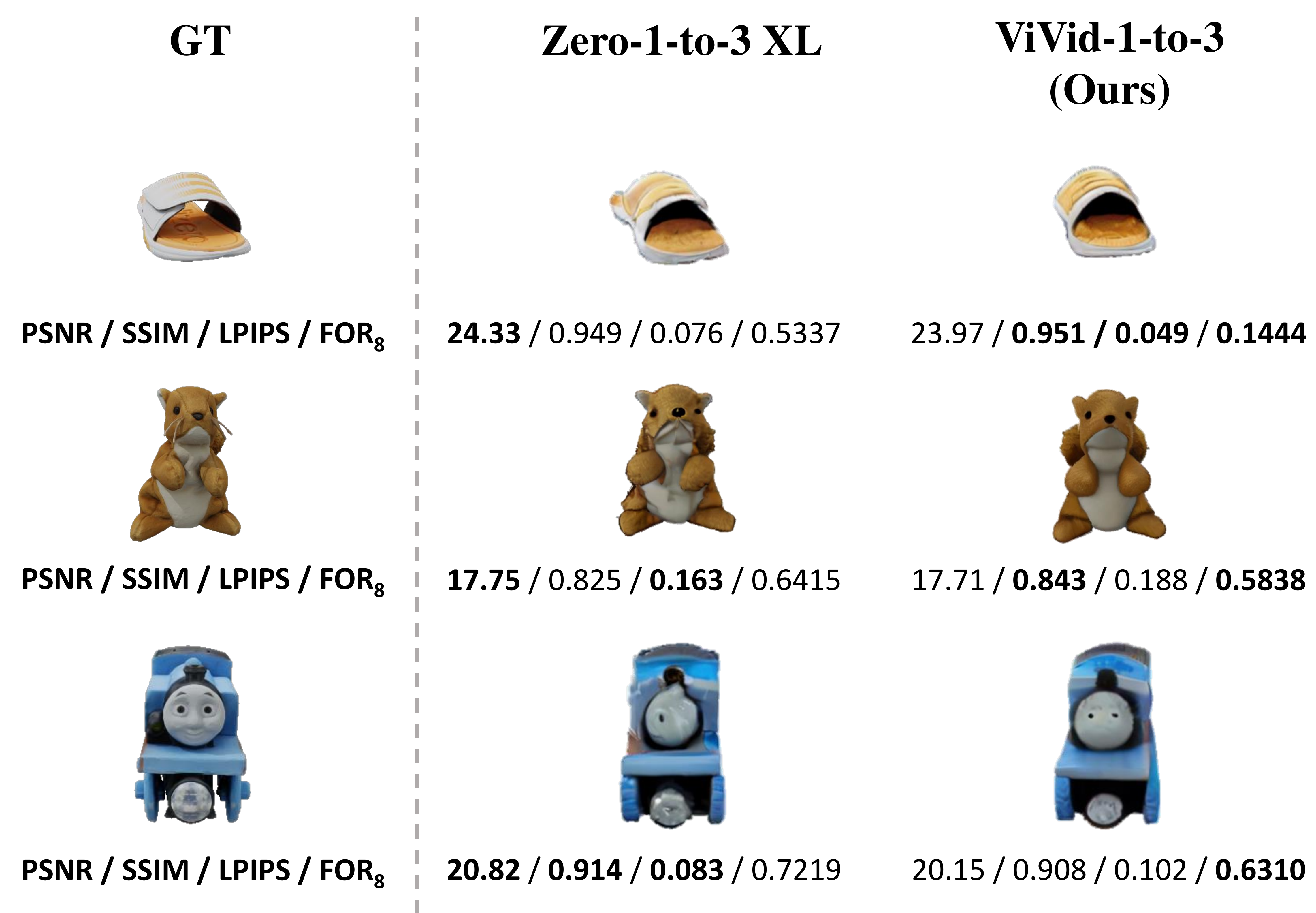}
    \caption{{\bf Shortcoming of standard image metrics -- }
    we show example \textbf{(left)} ground-truth renderings from the target view, and the novel-view images by \textbf{(middle)} Zero-1-to-3 XL and \textbf{(right)} our method.
    We show PSNR / SSIM / LPIPS / FOR$_8$ for each shape.
    As shown, standard image metrics---PSNR, SSIM, and LPIPS---may not correspond to how `accurate' each image is due to misalignments.
    Arguably, our renderings are more consistent and faithful to the ground truth.
    However, each metric provides a different story.
    We propose an optical flow-based metric, FOR$_k$, to compensate for this shortcoming and provide a holistic view.
    }
    \label{fig:metric}
\end{figure}

\paragraph{Datasets.}
To systematically evaluate the quality of the synthesized images, we rely on the 100 3D shapes from the Google Scanned Objects (GSO) dataset~\cite{downs2022google}.
Among the 1,030 objects in the dataset, we manually select 100 shapes that look interesting---for example, we ignore shapes such as cubes and spheres.
We render these images using the same lighting as in Zero-1-to-3~\cite{liu2023zero1to3}.
We render 25 views, with an elevation of 15 degrees and azimuth ranging from -45 to 45 degrees from manually selected views that capture the characteristics of the object.


\begin{figure*}
    \centering
    \includegraphics[width=\linewidth]{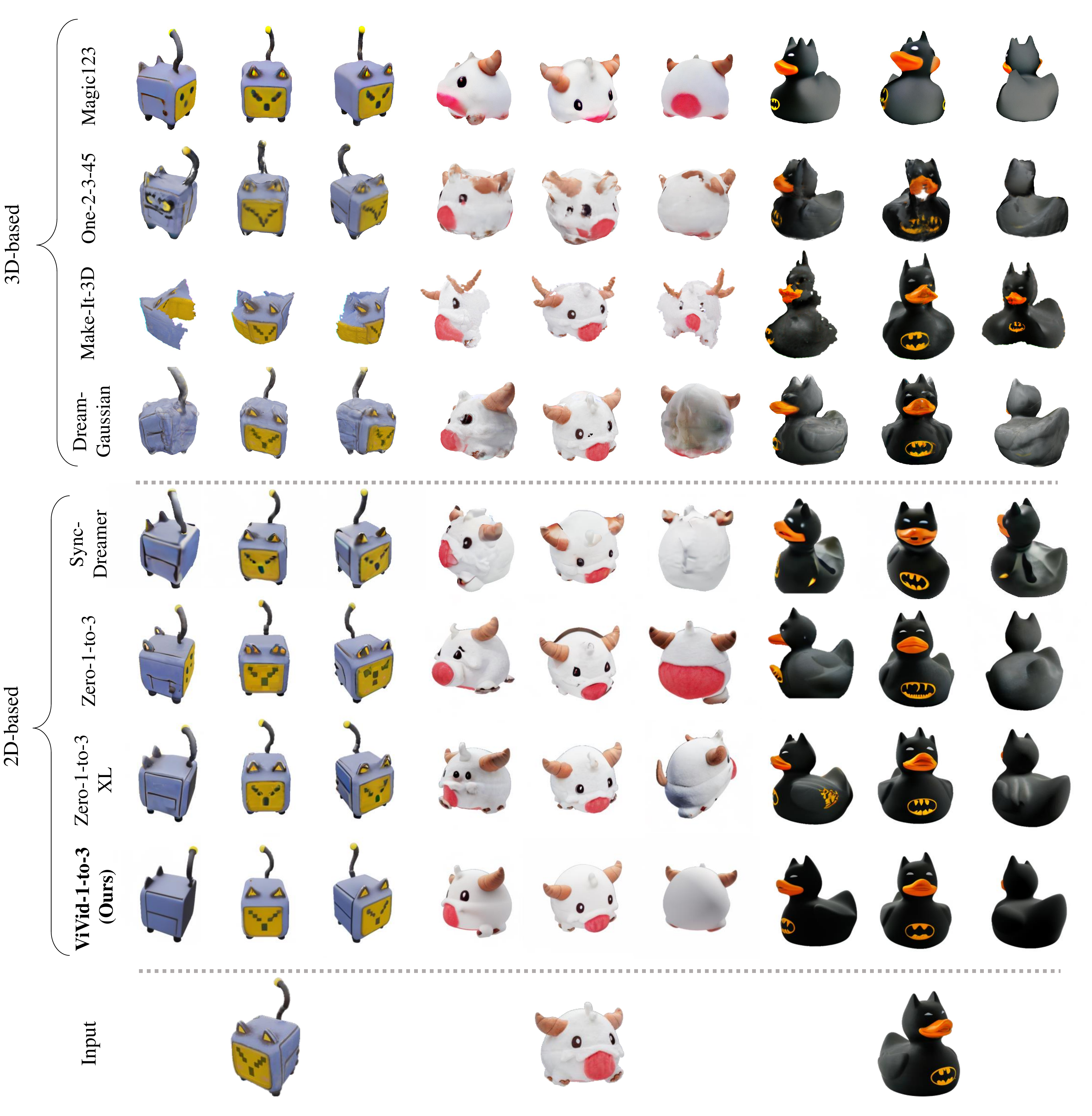}
    \caption{
    {\bf Qualitative highlights -- }
    we provide examples for multiple recent baselines.
    Our results are most consistent with the input image.
    Note the direction of the tail for the left object, the horns of the object in the middle, and the tail of the object on the right. 
    }
    \label{fig:qualitative}
\end{figure*}

\paragraph{Metrics.}
While results from diffusion-based novel-view synthesis models look good in general, this can sometimes be misleading as these synthesized images may not be of the desired view, and their contents may have changed as diffusion-based models often do not have the explicit notion of 3D space.
Thus, with the synthetic dataset (described above), we systematically measure the quality of the generated images.

While we also report standard images for novel-view reconstruction, Peak Signal-to-Noise Ratio (PSNR), Structural Similarity Index Metric (SSIM)~\cite{wang2004image}, and Learned Perceptual Image Patch Similarity (LPIPS)~\cite{zhang2018unreasonable}, we notice that these metrics do not faithfully resemble the synthesis quality.
It is because synthesizing a novel view from a single image is inherently an ill-posed problem, and even the slightest error in the implied 3D estimate can lead to a large error for the standard metric; see~\cref{fig:metric}.
This is expected, since these standard metrics are well known to be either highly sensitive to misalignment as in the case of PSNR and SSIM, or lead to measures that cannot distinguish between good and poor alignment due to insensitivity, \eg, with LPIPS.

\paragraph{Optical flow metric -- FOR$_k$.}
We thus propose to use a metric that measures how many of the rendered pixels are close enough to where they really should be -- optical flow.
To retrieve optical flow, given that the generated images and the ground truth are supposed to be similar, we use RAFT~\cite{teed2020raft}.
We then measure the ratio of flow estimates that deviate significantly (we use various thresholds) from RAFT estimates to account for the potential errors that RAFT itself may make.
We denote our metric as FOR$_k$, where $k$ is the pixel threshold---we use 8 and 16 as RAFT is reported~\cite{teed2020raft} to make an average pixel error of 5 pixels on the KITTI benchmark~\cite{geiger2012we}.
Note that our metric is similar to how the KITTI benchmark measures optical flow accuracy by counting the ratio of optical flow outliers.
Ideally, if the rendering was perfect, this metric would be zero, and the lower the ratio of flow outliers the more accurate.
Since we are looking at estimated optical flow, which relies on appearance, this metric takes into account both the faithfulness of the appearance and the alignment; see~\cref{fig:metric}.




\subsection{Qualitative comparison}

\vspace{-\customparskip}
\paragraph{Baselines.}
We compare our method against the following 2D and 3D baselines:
\begin{itemize}
    \item {\bf 2D) Zero-1-to-3 (XL)}~\cite{liu2023zero1to3,deitke2023objaversexl}:
    the base model for our novel-view diffusion model. 
    We compare against both the original version and the one fine-tuned on Objaverse-XL~\cite{deitke2023objaversexl}.
    \item {\bf 2D) SyncDreamer}~\cite{liu2023syncdreamer}:  
    a 2D novel-view diffusion model that improves the consistency of Zero-1-to-3 XL by finetuning it with pre-defined fixed views.
    For this method, we thus show the nearest view, in its pre-defined view set.
    \item {\bf 3D) Magic123}~\cite{abs-2306-17843}:
    a method that consists of two stages of coarse-to-fine generation process for textured 3D mesh.
    Firstly it utilizes both 2D~\cite{rombach2022high} and 3D~\cite{poole2022dreamfusion} diffusion prior to train NeRF. 
    It then converts the trained NeRF to DMTet~\cite{shen2021deep} to generate high-resolution 3D textured models.  
    \item {\bf 3D) Make-It-3D}~\cite{Tang_2023_ICCV}: 
    This method first uses the reference image and its monocular depth estimate to optimize a NeRF. 
    It then uses the Score Distillation Sampling (SDS)~\cite{poole2022dreamfusion} loss with a text prompt obtained via BLIP-2~\cite{li2023blip} and optimizes a textured point cloud, which is then rendered as novel views via deferred neural rendering.
    \item {\bf 3D) One-2-3-45}~\cite{liu2023one}: a method that trains a straightforward and efficient multi-view reconstruction model with a 3D convolutional neural network using images generated from Zero-1-to-3 (XL)~\cite{liu2023zero1to3,deitke2023objaversexl}. 
    \item {\bf 3D) DreamGaussian}~\cite{tang2023dreamgaussian}:
    a method that leverages the recently popularized 3D Gaussian Splatting~\cite{kerbl20233d} for efficient 3D textured mesh generation.

\end{itemize}
\vspace{1em}
For all baselines, we use the official code provided by the authors and use default parameters.
We visualize each baseline from a similar view, which we manually align for 3D methods due to the different coordinate conventions of each method, and the inconsistency between the pose estimates and the requested pose.

\paragraph{Discussion \cref{fig:qualitative}.}
We first demonstrate our results qualitatively in \cref{fig:qualitative}.
As shown, our method provides results that are much more consistent with the input image.
Note that many of the methods shown, qualitatively, look good. 
However, upon close inspection, these models are often not consistent with either the input image, especially as the target view deviates strongly from the input image.
For example, the tail direction of the left object, the horns of the middle object, or the ears and the tails of the right object.

We notice that the methods that utilize explicit 3D representations show worse quality than the pure 2D ones.
This is potentially due to their reliance on monocular depth estimates from off-the-shelf methods~\cite{ranftl2020towards,ranftl2021vision}, and the pose inconsistencies that 2D novel-view diffusion models bring.
While their renderings themselves are 3D consistent by construction, they thus tend to be blurrier and less faithful to the input image.
%

\begin{figure}
    \centering
    \includegraphics[width=\linewidth]{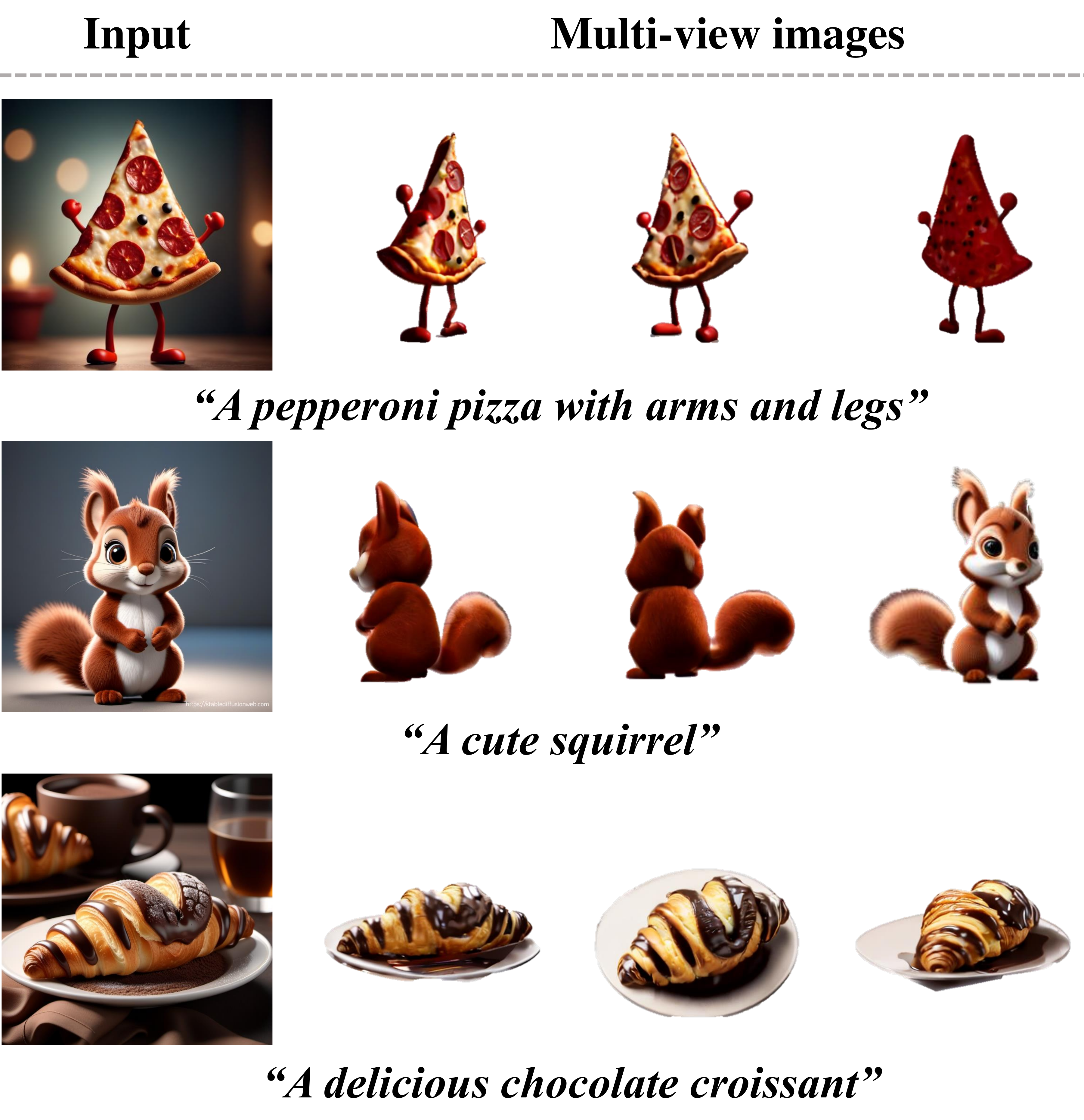}
    \caption{
    {\bf From text to novel-view synthesis --}
    we show examples of multiple views of objects being generated purely from text prompts via Stable Diffusion~\cite{rombach2022high} and our method.
    We show both the original view and the novel view.
    }%
    \label{fig:text-to-nvs}
\end{figure}
\paragraph{From text to novel views -- \cref{fig:text-to-nvs}.}
\label{sec:text-to-nvs}
We further demonstrate rendering multiple views of objects purely from text.
We generate an image of an object with Stable Diffusion, remove the background via \citet{ranftl2021vision}, and then use our method to generate novel views.
The generated views are not only consistent with the input view but also preserve the semantics of the input prompt. 
Our framework yields a high-quality text-to-novel view synthesis model when combined with the denoising pipeline of stable diffusion.
%


\begin{figure*}
    \centering
    \includegraphics[width=\linewidth]{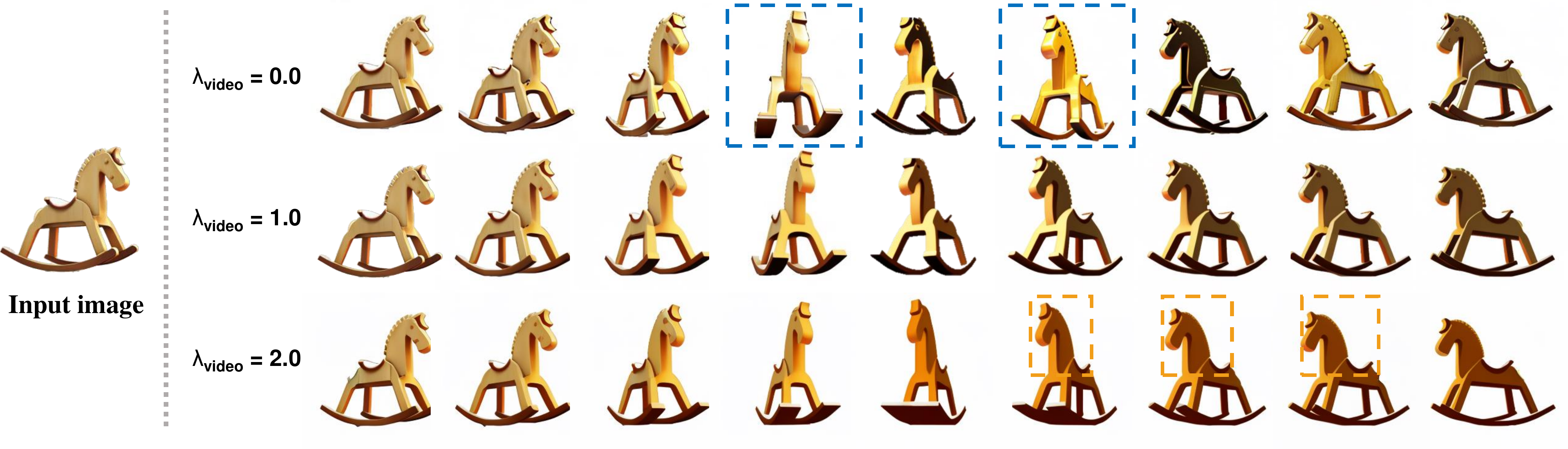}
  \caption{
  {\bf Effect of hyperparameters -- }
  we show an example of our method when \textbf{(top)} video diffusion is not used hence equivalent to Zero-1-to-3 XL~\cite{liu2023zero1to3,deitke2023objaversexl}, \textbf{(middle)} with optimal hyper parameters, and \textbf{(bottom)} with too much influence from video diffusion model.
  Without video diffusion the poses are inconsistent---they change abruptly from one image to another (marked with \textcolor{Blue}{blue boxes}).
  With too much video diffusion the content smooths out, losing detail (marked with \textcolor{Orange}{orange boxes}).
  }%
        \label{fig:hparam_qualitative}
\end{figure*}

\subsection{Quantitative results}

\paragraph{Baselines.}
Due to the limited amount of computational resources, we focus our efforts to peer-reviewed baselines at the time of writing: Zero-1-to-3 (XL)~\cite{liu2023zero1to3,deitke2023objaversexl} and Make-It-3D~\cite{Tang_2023_ICCV}.
As before, we use the official implementations, with their default configurations.
For Make-It-3D, we render images at $800\times800$ which is the default rendering resolution, and then resize it to $512\times512$ with bilinear sampling. 

\begin{figure}
    \centering
    \includegraphics[width=\linewidth, trim=14 363 460 97, clip]{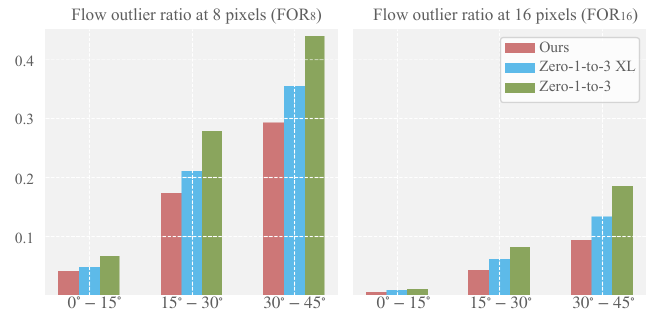}
    \caption{
    {\bf Optical flow outlier ratio -- }
    we report the optical flow outlier ratio with varying thresholds (8 and 16 pixels) for each method for novel view images generated for different viewing angles.
    Our method provides significant improvement over Zero-1-to-3 XL~\cite{liu2023zero1to3,deitke2023objaversexl} and outperforms all methods.
    }
    \label{fig:deg_vs_flow}
\end{figure}

\paragraph{Flow-based metric -- \cref{fig:deg_vs_flow}.}
We report the flow-based metric at varying thresholds---8 and 16 pixels---in \cref{fig:deg_vs_flow}.
Here, we exclude Make-It-3D as its results were significantly worse than others, as we will show via standard metrics in \cref{tab:quantitative_img}.
Our method significantly improves over Zero-1-to-3 XL, especially when the requested viewing angle is distant from the original view. 
This demonstrates how video diffusion helps in providing more consistent renderings.

\begin{table}
    \centering
    \setlength{\tabcolsep}{12pt}
    \resizebox{\linewidth}{!}{
    \begin{tabular}{l c c c}
    \toprule
         & PSNR$\uparrow$ & SSIM$\uparrow$ & LPIPS$\downarrow$ \\
    \midrule
        Make-It-3D~\cite{Tang_2023_ICCV} & 17.13 & 0.841 & 0.135 \\
        Zero-1-to-3~\cite{liu2023zero1to3} & 22.66 & 0.902 & 0.106 \\
        Zero-1-to-3 XL~\cite{liu2023zero1to3,deitke2023objaversexl} & 23.47 & 0.909 & 0.101 \\
        ViVid-1-to-3 (Ours) & \bf{24.05} & \bf{0.917} & \bf{0.099} \\
    \bottomrule
    \end{tabular}
    }
    \caption{{\bf Image metrics -- }
    We report the PSNR, SSIM, and LPIPS for each method for all views.
    Our method performs best.
    }%

    \label{tab:quantitative_img}
\end{table}
\paragraph{Image-based metrics --\cref{tab:quantitative_img}.}
For completeness, we further report the standard image-based metrics in \cref{tab:quantitative_img}.
Our method also improves over the state of the art in terms of traditional image quality metrics.
Interestingly, Make-It-3D, by focusing on building an explicit 3D representation, loses quality when it comes to actual 2D renderings.






\label{sec:ablation}

\begin{table}
    \centering
    \resizebox{\linewidth}{!}{
    \begin{tabular}{@{}c c c c c c c@{}}
    \toprule
                $\hparam{video}^s$& $\hparam{video}^e$& PSNR$\uparrow$ & SSIM$\uparrow$ & LPIPS$\downarrow$ & 
                FOR$_8$$\downarrow$ & FOR$_{16}$$\downarrow$\\ 
        \midrule
        1.0 & 0.5 & \bf{22.55} & \bf{0.905} & \bf{0.105}  & \bf{0.2923} & \bf{0.0944}\\
        1.0 & 1.0 & 22.55 & 0.905 & 0.106 & 0.2958 & 0.0949\\
        1.0 & 0.0 & 22.48 & 0.905 & 0.105 & 0.3082 & 0.0995\\
        1.5 & 0.5 & 22.44 & 0.904 & 0.106 & 0.2952 & 0.1072\\
        \bottomrule
    \end{tabular} 
    }
    \caption{
    {\bf Effect of hyperparameters -- }
    we show all metrics for the azimuth range of 30--45 degrees for various scheduling of $\hparam{view}$ and $\hparam{video}$. 
    We keep $\hparam{view}{=}1$ and schedule $\hparam{video}$, where we denote the linear scheduling as $\hparam{video}^s$-$\hparam{video}^e$-$t$, where $\hparam{video}^s$ is the starting $\hparam{video}$ value, $\hparam{video}^e$ is the end value at timestep 50.
    }%
    \label{tab:hparam_psnr}
\end{table}

\paragraph{Effect of hyperparameters $\hparam{view}$ and $\hparam{video}$ -- \cref{tab:hparam_psnr} and \cref{fig:hparam_qualitative}.}
As discussed earlier in \cref{sec:main_method}, the choice of $\hparam{view}$ and $\hparam{video}$ matters.
We investigate multiple parameter settings and report a subset of our search in \cref{tab:hparam_psnr}.
Note that our optical flow-based metric, FOR$_k$ is more distinctive.
As shown, relying either too much or too little on the video diffusion model is suboptimal.
A representative example of both cases is shown in \cref{fig:hparam_qualitative}.



\section{Conclusion}
\label{sec:conclusion}

We have presented a framework for novel-view synthesis, that poses the problem as a video generation problem, which allows combining novel-view diffusion models with video diffusion models.
By utilizing the strong priors learned within video diffusion models, we achieve more consistent novel-view synthesis results.
To compensate for the shortcomings of standard image-based metrics, we propose a novel metric based on optical flow.
We compared our method existing methods, achieving the state of the art.

\paragraph{Limitations and future work.}
While our method delivers improved consistency with the input image, it still does not have an explicit 3D model and can be multi-view inconsistent.
We note, however, that our method is complementary to other methods that focus on consistent novel-view synthesis including those that embed 3D models.
With our improved consistency in a pure 2D pipeline, a promising future direction would be to incorporate explicit 3D pipelines.

\clearpage
\paragraph{Acknowlegment.}

This work was supported in part by the MOTIE (Ministry of Trade, Industry, and Energy) in Korea, under Human Resource Development Program for Industrial Innovation (Global) (P0017311) supervised by the Korea Institute for Advancement of Technology (KIAT), the Natural Sciences and Engineering Research Council of Canada (NSERC) Discovery Grant, Digital Research Alliance of Canada, and Advanced Research Computing at the University of British Columbia.
{
    \small
    \bibliographystyle{ieeenat_fullname}
    \bibliography{main}
}

\clearpage
\maketitlesupplementary


\begin{figure*}
    \centering
    \includegraphics[width=\linewidth]{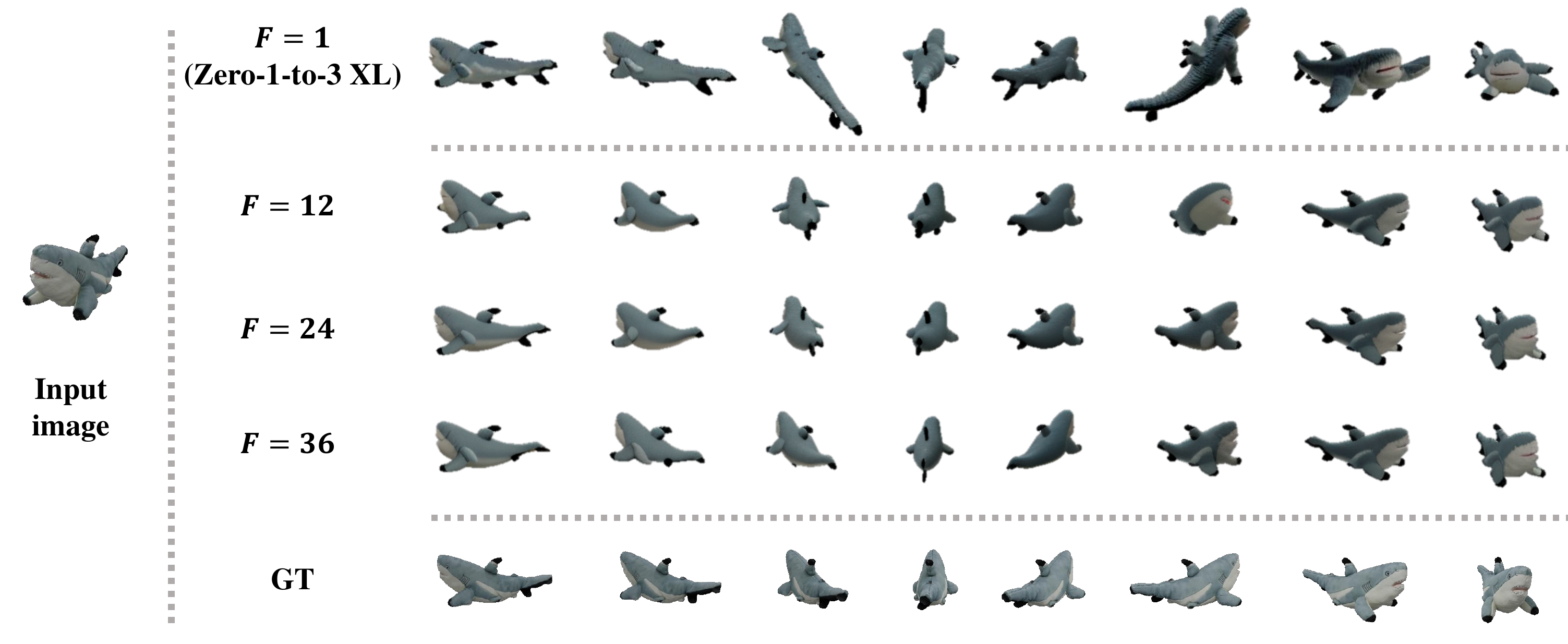}
    \caption{
    {\bf Effect of number of frames -- } We present an example of our framework by varying the number of frames. 
    }
    \label{fig:ablation_figure_framenum}
\end{figure*}

\begin{figure*}
    \centering
    \includegraphics[width=\linewidth]{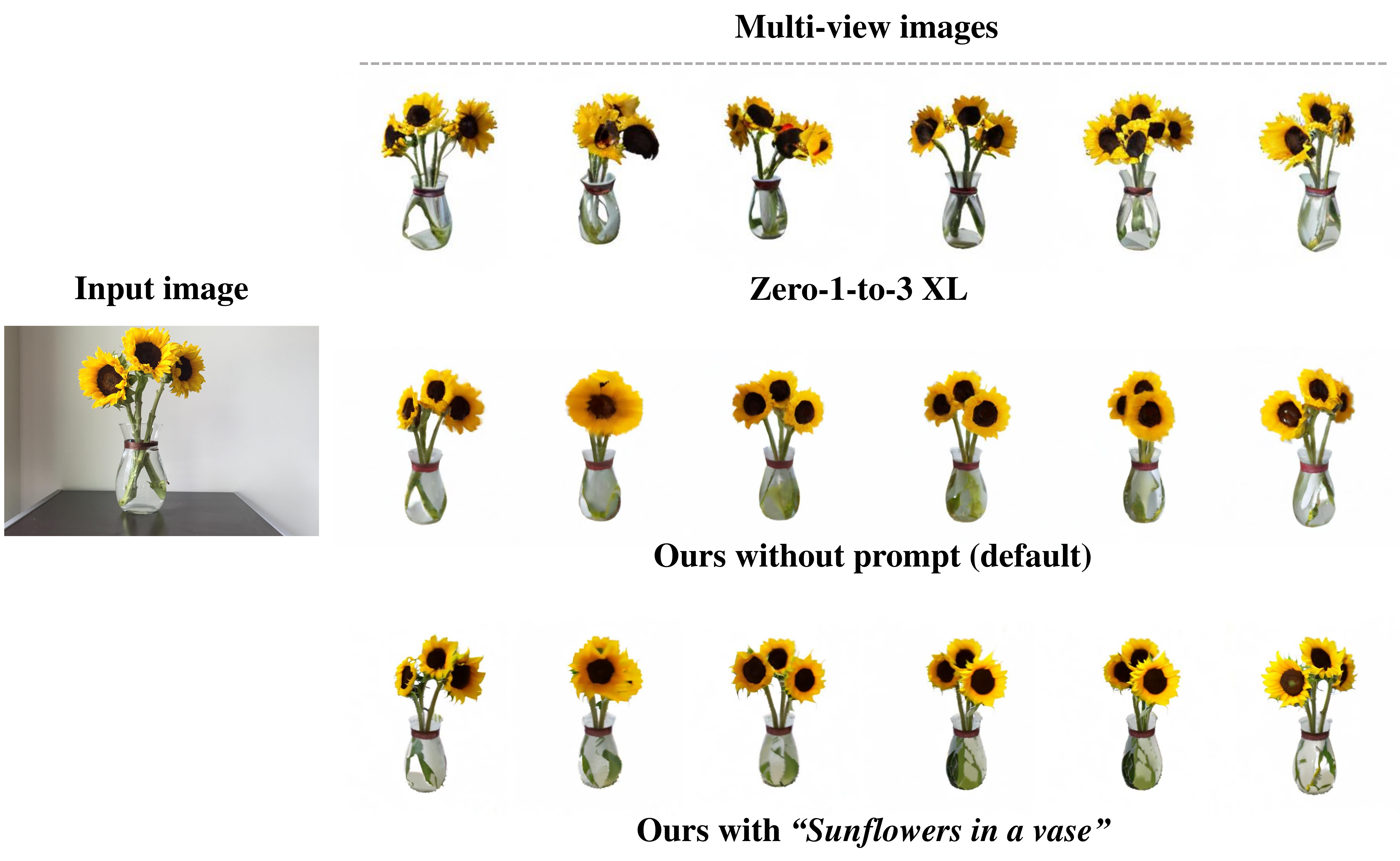}
    \caption{
       {\bf Effect of prompting -- }We show the effect of prompting on novel-view synthesis with our approach.
    }
    \label{fig:suppl_prompting}
\end{figure*}
This supplementary document includes additional content not covered in the main paper. We begin with the implementation details of our framework and then discuss the effectiveness of the design choices. Finally, we provide additional visual samples.

\section{Implementation}
For novel-view diffusion and video diffusion, we utilize the pre-trained Zero-1-to-3 XL~\cite{liu2023zero1to3,deitke2023objaversexl} and Zeroscope v2 576w~\cite{zeroscope_576}, respectively. 
The resolution of the rendered frames presented in the paper is $256^2$, but direct upscaling can be performed using Zeroscope v2 XL~\cite{zeroscope}. 
For the denoising scheduler, we employ the DPM solver~\cite{lu2022dpm} for both diffusion models instead of DDIM~\cite{song2020denoising} and conduct 50 inference steps. 
For $360\degree$ videos, we set the number of frame $F$ to 24, i.e., 24 frames video. The impact of the number of frames is discussed in \cref{sec:discussion}. 
Since we nullify the text prompt conditioning for the video diffusion model, as explained in the main paper, the classifier-free guidance is not applied, and the guidance scale for novel-view diffusion is set to 3.0.

\label{sec:implementation}

\section{Discussion}
\label{sec:discussion}
\paragraph{Effect of number of frames.}  
 As our framework incorporates video diffusion, we have the flexibility to modify the number of video frame $F$ in \cref{eq:our_denoising}. 
As shown in \cref{fig:ablation_figure_framenum}, we vary the number of frames to check the $360\degree$ rendered images. 
When each frame is generated independently (equivalent to Zero-1-to-3-XL~\cite{liu2023zero1to3,deitke2023objaversexl}), the synthesized images lack consistency and exhibit significant deviations from the ground truth samples. 
Leveraging the video diffusion prior considerably enhances the pose and shape consistency of the generated images, with accuracy improving as more frames are utilized. 
Therefore, users can adjust the number of frames based on the trade-off between their computational resources and the desired level of consistency.


\paragraph{Prompting strategy.}
Although we set the prompt for our video diffusion process as a null, i.e., $y=\O$ in \cref{eq:video_diffusion}, it does not mean that prompting is not possible. 
We found that in some cases, providing prompt can enhance the quality of our method.
As shown in \cref{fig:suppl_prompting}, the text conditioning \texttt{sunflowers in a vase} yields images of higher quality with better object-level details and hence improves the quality of novel-view synthesis.
This example further highlights the potential use cases of our method for high-resolution and editable novel-view rendering.

\section{Additional samples}

\begin{figure*}
    \centering
    \includegraphics[width=\linewidth]{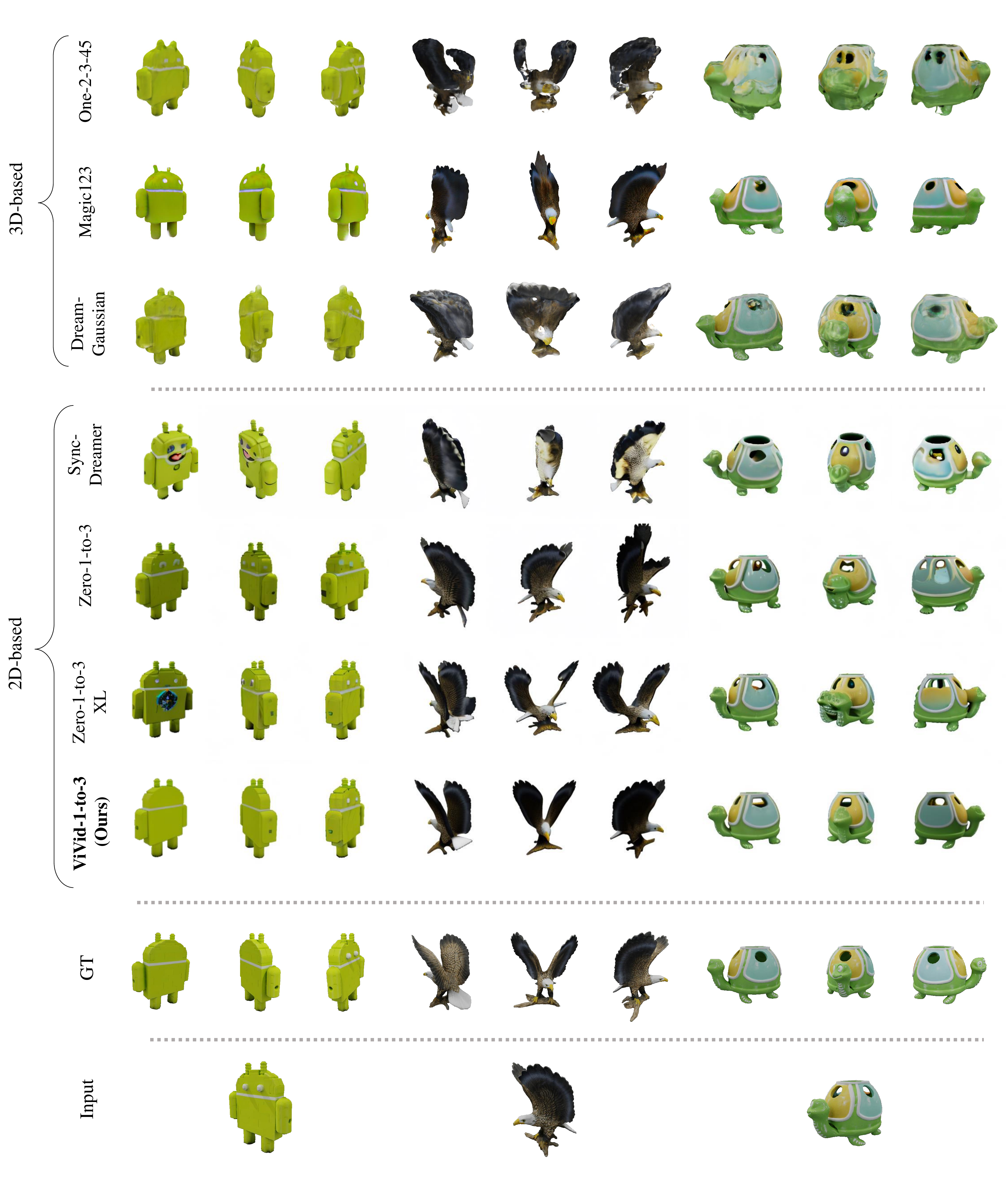}
    \caption{{\bf Additional qualitative results -- }
       Additional novel-view generation samples from GSO~\cite{downs2022google}, and comparison with 2D and 3D methods.
    }
    \label{fig:suppl_qual_vs_all}
\end{figure*}

\begin{figure*}
    \centering
    \includegraphics[width=\linewidth]{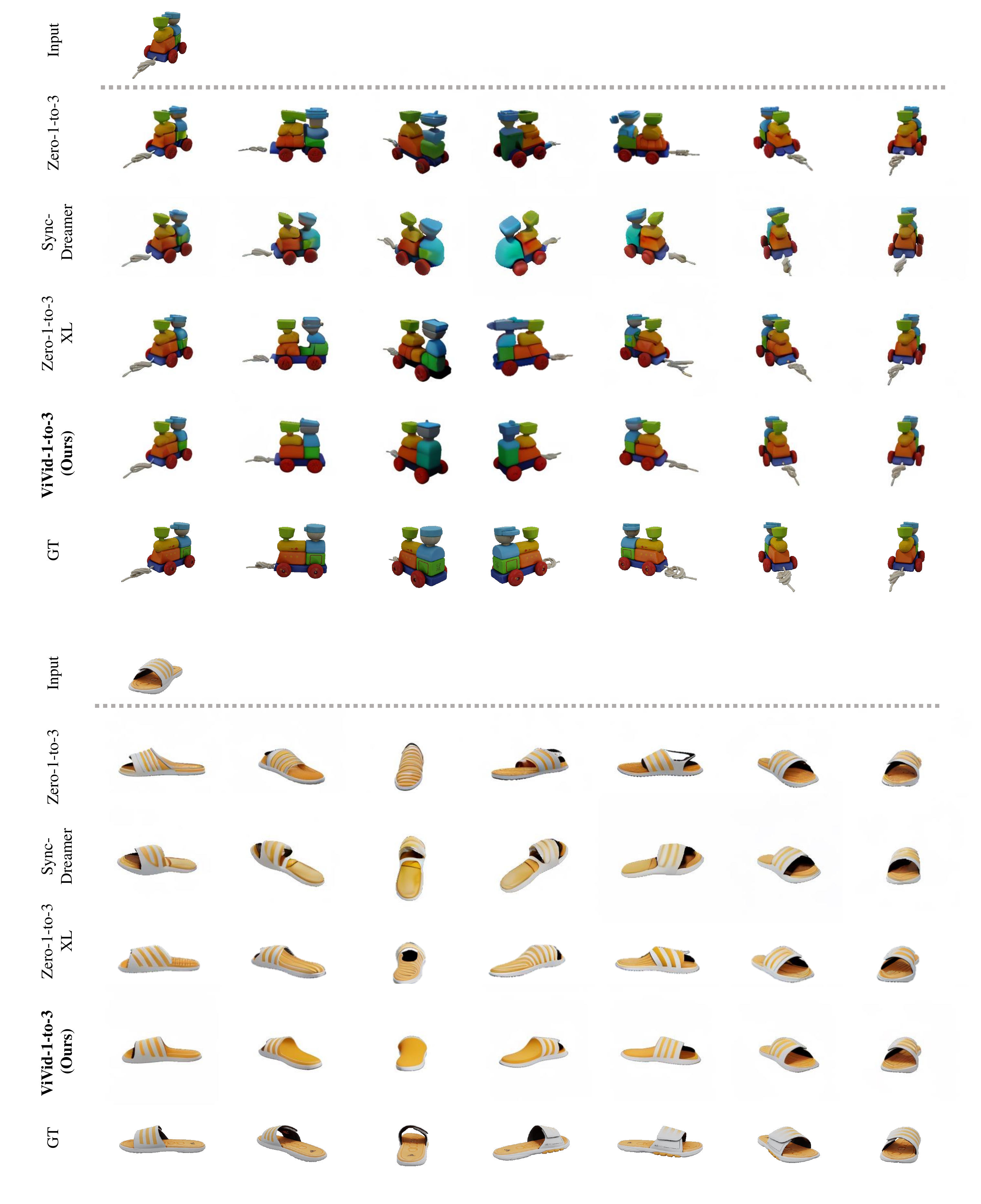}
    \caption{ {\bf Comparison to 2D methods -- }
       Additional multi-view synthesis samples of 2D-based methods on GSO~\cite{downs2022google} dataset.
    }
    \label{fig:suppl_qual_vs_2d}
\end{figure*}

\paragraph{Additional multi-view samples.}
 In \cref{fig:suppl_qual_vs_all} and \cref{fig:suppl_qual_vs_2d}, we offer additional samples of multi-view synthesis. 
 Here, we leverage the GSO dataset~\cite{downs2022google} to facilitate comparisons with ground truth samples. 
 In \cref{fig:suppl_qual_vs_all}, we compare our model to compare our model with 2D~\cite{liu2023zero1to3,deitke2023objaversexl,liu2023syncdreamer} and 3D methods~\cite{liu2023one,tang2023dreamgaussian,abs-2306-17843} in the same way as \cref{fig:qualitative} in the main paper.
 We additionally present multi-view samples of 2D-based methods~\cite{liu2023zero1to3,deitke2023objaversexl,liu2023syncdreamer} including ours in \cref{fig:suppl_qual_vs_2d}, to verify multi-view consistency and visual quality of the competitive models. 









\end{document}